\documentclass[conference]{IEEEtran}
\IEEEoverridecommandlockouts
\usepackage{cite}
\usepackage{amsmath,amssymb,amsfonts}
\usepackage{algorithmic}
\usepackage{graphicx}
\usepackage{tabularx}
\usepackage{textcomp}
\usepackage{xcolor}
\usepackage{booktabs}
\usepackage{stfloats} 
\usepackage{xcolor}
\usepackage{colortbl}
\def\BibTeX{{\rm B\kern-.05em{\sc i\kern-.025em b}\kern-.08em
    T\kern-.1667em\lower.7ex\hbox{E}\kern-.125emX}}

\makeatletter
\newcommand{\linebreakand}{%
  \end{@IEEEauthorhalign}
  \hfill\mbox{}\par
  \mbox{}\hfill\begin{@IEEEauthorhalign}
}
\makeatother
\begin{document}

\title{SAR4SLPs: An Asynchronous Survey of Speech-Language Pathologists' Perspectives on Socially Assistive Robots}

\author{ Denielle Oliva$^1$, Abbie Olszewski$^2$, David Feil-Seifer$^3$
\thanks{$^1$Denielle Oliva, $^2$ Abbie Olszewski, and $^3$ David Feil-Seifer are with the Department of Computer Science and Engineering, University of Nevada, Reno, 1664 N. Virginia Street, Reno, NV 89557-0171, USA \emph{denielleo@unr.edu}, \emph{aolszewski@unr.edu} and \emph{dfseifer@unr.edu}}}

\maketitle

\begin{abstract}
Socially Assistive Robots (SARs) offer unique opportunities within speech language pathology (SLP) education and practice by supporting interactive interventions for children with communication disorders. This paper explores the implementation of SAR4SLPs (Socially Assistive Robots for Speech-Language Pathologists) to investigate aspects such as engagement, therapeutic strategy discipline, and consistent intervention support. We assessed the current application of technology to clinical and educational settings, especially with respect to how SLPs might use SAR in their therapeutic work. An asynchronous remote community (ARC) collaborated with a cohort of practicing SLPs to consider the feasibility, potential effectiveness, and anticipated challenges with implementing SARs in day-to-day interventions and as practice facilitators. We focus in particular on the expressive functionality of SARs, modeling a foundational strategy that SLPs employ across various intervention targets. This paper highlights clinician-driven insights and design implications for developing SARs that support specific treatment goals through collaborative and iterative design.

\end{abstract}


\section{Introduction}
Speech-language pathologists (SLPs) provide therapeutic services to over 3 million children in educational settings with speech, language, or communication disabilities in United States~\cite{ASHAworksettings}.The majority of SLPs work in schools and private practice; however, the demand for services often surpasses the existing workforce's capacity. This shortage is most acutely felt in under-served regions, such as rural areas and poor districts~\cite{SLPshortageSquires}. Many SLPs manage high case loads, which can delay or limit access to early intervention--an essential factor in successful treatment outcomes~\cite{TheraPlatformshortage}.

SLPs address many different clinical goals during intervention, working with those suffering with speech sound disorders, apraxia of speech, oral and written language disorders. Treatment for these disorders require explicit instruction and repetitions. For example, apraxia of speech is a neurological condition affecting motor planning for speech~\cite{childhoodapraxia}. In one of the treatment models, SLPs teach movement associated with verbal articulation by modeling facial expressions and subtle nonverbal cues. Furthermore, tactile stimulation techniques stress the affective dimension of accurate speech movement~\cite{childhoodapraxia}. Translating this nuanced, expressive communication into SAR-mediated therapy presents unique challenges.

Integrating additional tools as co-facilitators or external practice aides like socially assistive robots (SARs) may also help to facilitate therapy processes. SARs would alleviate some of the burden of individual professionals, thereby also increasing individual bandwidth of SLPs and increasing practice opportunities. Research indicates that children interact with SARs similar to how they interact with their peers. For children with communication disorders, SARs can offer a lower-risk, less intimidating interaction model~\cite{Smakman2022-dc}. This quality makes SARs appealing, aiming to serve as a capable therapy multipliers as companions, practice partners, etc. 

We conducted a mixed methods study through an asynchronous remote community, fourteen SLPs in the state of Nevada over four weeks. Through this ARC study, we investigated the insights of SLPs on SARs, how facial expressions play a role in their practice, and appropriate design considerations for an assistive robot in varying roles during intervention. Results of the study suggest promising and positive outcomes for the use of SARs as tools in speech-language therapy.  The results of this study helps inform the use of SARs for SLP interventions, ensuring that implementations of SARs are successful and accepted by both clients and therapists.

\section{Related Work}

Using SARs as tools in therapy has shown promising results, particularly in autism spectrum disorder (ASD) therapy which often overlaps with the scope of SLP interventions. Feil-Seifer et al.~\cite{feil-seifer2005defining} and Scasalletti et al.~\cite{scassellati2012robots} laid foundational work in integrating SARs with therapeutic roles, supporting children with ASD and other communication disorders. However, these prior works do not take a user-centered approach when implementing a SAR in a clinical space. SARs have also proved to be a good vehicle for practice outside of traditional therapy spaces. In Shi et. al SARs were evaluated for a long-term tutoring role as well as their performance as autonomous affect-capable systems~\cite{shi2021personalizedaffectawaresociallyassistive}. The premise of using SARs for co-facilitators or activity partners can be translated to behaviors targeted in SLP interventions. Spitale et al. take this into consideration, using a SAR during an activity targeting language impairments~\cite{spitale2023using}.

Nonverbal communication, including facial expressions, used by SLPs serve a foundational component for many existing intervention strategies with target behaviors ranging from articulation to social skills. To address facial expressions during intervention, a SAR must be designed with expressive functionality in consideration. This work investigates the role of expression in various therapy applications in SLP spaces in order to make adaptable and dynamic SAR behaviors. 

Within the scope of speech-language pathology, nonverbal expressiveness and affect interpretation can be found intertwined within many of their practices \cite{friedman1980understanding, buck1972communication}. Camras et al. and Frith et al. provide foundations with socialization processes through which children develop interpretation and expression of emotions \cite{camras1985socialization, chris2009role}. These concepts can be aligned with from Cowen and Keltner's taxonomy of 28 emotional states displayed in naturalistic expression, providing a framework for how we define emotions in this study~\cite{Cowen2019-xx}.

In terms of applying expression to SARs, research by Saerback et al, Kennedy et al, and Kalegina et al illustrate its crucial role in developing user perceptions and positive therapy outcome~\cite{saerbeck2010, kennedy2017impact, kalegina2018}. In this work we identify relevant design considerations for developing an expression capable SAR. Through co-designs with applied communities, this methodology can bridge the gap between technical development and real-world application, making these systems more valuable and usable to the clinicians that they aim to support.

Asynchronous Resource Community (ARC), is a research method that uses the flexibility of online platforms to facilitate discussions asynchronously and connect participants in isolated forum settings~\cite{10.5555/3021319.3021320}. We conducted an ARC because it aligns with the demanding and varied availability of SLPs, allowing them to participate on their own time. This methodology also allows us to connect participants from across geographic locations. This ARC connected participants from varying regions for their unique experiences~\cite{10.1145/3479546}.  Through an ARC researchers can prompt participants for information and task completion that is accessible and low risk~\cite{Bhattacharya2021-cz}. 

\section{Methods}

\begin{table*}[t]
\caption{This table outlines the activities that participants completed over four weeks. \\Qualitative activities are denoted with \{QL\} and quantitative questions are denoted with \{QT\}. }
\label{tab:activity_overview}
\resizebox{\textwidth}{!}{%
\begin{tabular}{@{}ccll@{}}
\toprule
\textbf{Week \#} & \textbf{Weekly Focus} & \textbf{Activities}  & \textbf{Research Questions} \\ \midrule

1 & \textbf{Introductions}                    & \begin{tabular}[c]{@{}l@{}}Technology Literacy and Experience\\ - How often do you use technology? \textbf{\{QT\}}\\ - How would you rate your tech literacy? \textbf{\{QT\}}\\ - Tools and technology you use in intervention \textbf{\{QL\}}\end{tabular}  & \textit{\textbf{\begin{tabular}[c]{@{}l@{}}RQ1. What initial insight can \\ introductions and experience \\ provide?\end{tabular}}} \\

\rowcolor[HTML]{C0C0C0} 
2 & \textbf{\begin{tabular}{c}Expressions and\\ Legacy Practices\end{tabular}} & \begin{tabular}[c]{@{}l@{}}What does your day to day look like?\\ - complete a survey with questions about existing\\ tools that you use \textbf{\{QL\}}\\ - If you could add a SAR to your intervention\\ practices, what activities could be enriched with \\ a SAR \textbf{\{QL\}}\\ \\ How do expressions play a part in your therapy \\ interventions? \\ - Using the 28 emotional states, classify what \\ expressions you use and how \textbf{\{QL\}} \textbf{\{QT\}}\\ - How do you implicitly and explicitly use \\ expressions in your therapy strategies? \textbf{\{QL\}}\end{tabular} & \textit{\textbf{\begin{tabular}[c]{@{}l@{}}RQ1. Detailed responses of an SLPs\\ day to day strategies and tools can \\ show how SARs can be implemented\\ \\ RQ2. Using the 28 emotional states, \\ expressiveness can be translated to \\ actionable tasks on implementing a\\ SAR\end{tabular}}} \\

3 & \textbf{Roles in Therapy}                 & \begin{tabular}[c]{@{}l@{}}What roles can SARs play in different therapy\\ spaces? \textbf{\{QL\}} \textbf{\{QT\}}\\ - In-person therapy interventions\\ - At home practice activities\\ \\ What benefits do you anticipate from using\\ SARs in therapy practices? \textbf{\{QL\}}\end{tabular}  & \textit{\textbf{\begin{tabular}[c]{@{}l@{}}RQ3. From their existing strategies and\\ existing roles, how do SARs fit within\\ those constraints\\ \\ RQ4/RQ5. Acceptance metrics and designs;\\ Role descriptions and justifications\end{tabular}}}                                                \\

\rowcolor[HTML]{C0C0C0} 
4 & \textbf{Design Requirements}  & \begin{tabular}[c]{@{}l@{}}Using varying target behaviors, what robot \\ design features would be helpful\\ - Explain your choices \textbf{\{QL\}}\end{tabular}  & \textit{\textbf{\begin{tabular}[c]{@{}l@{}}RQ4/RQ5. Use cases and implementation\\ features for varying behaviors and therapy\\ strategies\end{tabular}}}  \\ \bottomrule
\end{tabular}%
}
\end{table*}

This work focuses on identifying opportunities for implementing an effective and human-centered SAR through appropriate design considerations. We build on existing SAR implementations through a human centered design perspective catered for SLPs. Our research questions are:

\begin{itemize}
    \item \textbf{RQ1} What experiences from an SLP's typical routine can inform the design of a SAR?
    \item \textbf{RQ2} How does an SLP encourage engagement through non-verbal communication?
    \item \textbf{RQ3} How do the experiences of SLPs influence their opinions and willingness to use a SAR during interventions?
    \item \textbf{RQ4} What is required for the acceptance and usability of a SAR agent during an intervention?
    \item \textbf{RQ5} What development considerations would be most helpful based on the preferences of SLPs?
\end{itemize}

To answer these questions, we employ an ARC focusing on the insights of SLPs through reflective and asynchronous discussions. This method allowed us to find context-rich responses from SLPs actively engaged in their individual practices. By integrating their lived experiences and strategies, the ARC offers a human-centered framework for finding design requirements that are not only functional, but meaningfully aligned with the nuances of real-world intervention.
 
\subsection{Participants and Recruitment}
The ARC conducted recruited 23 licensed SLP professionals in the state of Nevada\footnote{This study was reviewed by the Institutional Review Board (IRBNet ID: 2215653) of the University of Nevada, Reno and classified as exempt.}. Participants were recruited through an interest survey sent to online Nevada SLP communities. To be eligible for this study, participants needed to be a licensed speech language pathologist in Nevada and serve clients in K-12 education. Twenty-three participants completed the initial interest survey and were invited to join the private online Slack community. 13 out of the original 23 participants completed on-boarding instructions for the SAR4SLPs online community. All participants had the Certificate of Clinical Competence in Speech Language Pathology (CCC-SLP)~\cite{ashaInformationAbout}. Participant experience ranged from one year of experience to 33 years of experience with the average years of experience being around 13.4 years. The study included two males and 11 females.

We used Slack to operate our ARC~\cite{Gofine2017-vw}. Using this platform allowed for asynchronous discussions, anonymity to be maintained through pseudonyms, and was accessible on multiple devices. This platform also allowed the research team to easily share any files or formatted messages for informational posts/discussion prompts. Discussion posts were mostly reflective questions to contrast the guided responses required in the survey questions. Throughout the discussions, the research team provided icebreakers or contributed to the conversations to encourage discussions to start among participants. 

\subsection{Activity Prompts}
We employed a mixed methods study design, asking both quantitative questions and open-ended questions. Over the course of four weeks, participants were asked to complete surveys and respond to discussion questions. The activities, and the objectives they are tied to, are detailed in Table~\ref{tab:activity_overview}. Each activity was designed to take approximately 10-20 minutes each to accommodate the schedules of the participants. The activities asked for recall and generative responses similar to previous ARC studies~\cite{10.1145/3663384.3663387}.

Four members of the research team were present in the Slack community to moderate and to monitor conversations to answer any questions that the participants had. The compensation for each participant was $\$50$ per activity completed. The total possible compensation package was $\$350$. The activities were designed based on a weekly focus aimed at answering related research questions. These questions and weekly focus are shown in Table~\ref{tab:activity_overview}. 

\begin{table*}[]
\caption{This table outlines the discussion point that participants responded to over four weeks.}
\label{tab:discussion_overview}
\resizebox{\textwidth}{!}{%
\begin{tabular}{@{}ccll@{}}
\toprule
\textbf{Week \#} & \textbf{Weekly Focus} & \textbf{Discussion Points}  & \textbf{Channels} \\ \midrule

1& \textbf{Introductions}                    & \begin{tabular}[c]{@{}l@{}}Introducing yourself to the group\\ - Pseudonym \\ - How long have you been practicing\\ - What areas do you specialize/work in\\ - An interest that you have outside of your work\end{tabular}                                                                                                                                                       & \textit{\textbf{General}}          \\
\rowcolor[HTML]{C0C0C0} 
2& \textbf{Expressions and Legacy Practices} & \begin{tabular}[c]{@{}l@{}}Do you think certain emotional states are more \\ difficult to express or interpret through facial \\ expressions? How might cultural differences, \\ individual variability, or context influence the \\ way these emotions are conveyed and understood \\ in speech-language therapy?\end{tabular}                                                  & \textit{\textbf{Robot Discussion}} \\
3& \textbf{Roles in Therapy}                 & \begin{tabular}[c]{@{}l@{}}How might Socially Assistive Robots (SAR) be \\ integrated into SLP interventions to support \\ different aspects of therapy, such as articulation, \\ language development, or social communication? \\ What factors should be  considered when \\ determining their effectiveness and suitability for \\ different client populations?\end{tabular} & \textit{\textbf{Robot Discussion}} \\
\rowcolor[HTML]{C0C0C0} 
4& \textbf{Design Requirements}              & \begin{tabular}[c]{@{}l@{}}Based on what we've explored over the past four \\ weeks, what do you think are the most essential \\ design features for social robots to effectively \\ support specific target behaviors? Are there any \\ features you believe are currently overlooked in \\ human-robot interaction?\end{tabular}                                               & \textit{\textbf{Robot Discussion}} \\ \bottomrule
\end{tabular}%
}
\end{table*}

\subsection{Discussion Questions}
Participants were asked to respond to reflective questions based on the focus of the week's content. Prompts were designed to encourage discussion among the participants and to gather information to supplement the survey questions. These questions are summarized in Table~\ref{tab:discussion_overview}. These encouraged participants to respond to insights of other SLPs as well as provide opposing views and additive insights. 

\subsection{Thematic Analysis}
A thematic analysis (TA) was conducted to examine the qualitative data collected through the weekly discussion prompts and the open ended questions included in the activities. TA allowed us to identify recurring patterns, concerns, and priorities expressed by the participants. We followed Braun and Clarke's~\cite{Braun01012006} six phase framework for TA. Initial codes were generated to encapsulate features of the data which were then organized into candidate themes. Themes were refined through comparison and cross-referencing with the original transcripts to maintain accuracy of the participant's perspectives. This approach enabled us to map the themes to the broader goals for design considerations of a SAR. 

\section{Results}
This section reports the themes found among the responses in the activities and surveys in the prior section. 13 participants consistently completed all of the activities. Clinician perspectives provide design considerations contributing to future implementation of SARs in SLP spaces. 

\subsection{Technology Literacy - Week 1}
\subsubsection{Activity}
Participants were asked to report on their familiarity and current usage of technology in their daily interventions. A total of 13 participants out of 14 completed Week 1's survey. 
Participants reported varying levels of technology use during intervention sessions. 8\% (n=1) indicated to rarely use computers or tablets, 46\% (n=6) reported to sometimes use computers and tablets, 31\% (n=4) often use computers and tablets, and 15\% (n=2) always use technology in their therapy sessions. 

Participants were also asked to report on their previous knowledge of socially assistive robots. The majority of participants 69\% stated that they had \textit{never heard of SARs}, 23\% \textit{had heard some information}, and only one participants had heard or talked about SARs prior to the ARC. 

Participants reported using a wide range of external tools during their sessions. These include traditional toys, paper-based activities/worksheets, Augmentative and Alternative Communication (AAC) devices, digital applications, and sensory integration materials.

\subsection{Clinical Application/Nonverbal Communication - Week 2}

\subsubsection{Activity}
Participants detailed individual goals for their practice and described their use of facial expressions and other nonverbal cues. This was divided into two activities. We asked participants to pick a specific client and describe their goals, strategies, and interactions during their programs. Responses included goals like expressive language and speech sound production, emotional regulation, AAC use, and pragmatic language skills. Intervention approaches were often curated for individual clients, combining structured and child-led methods. Tools used varied from low-tech materials like books to high-tech AAC devices and digital therapy applications. 

 Week 2 also focused on nonverbal communication applications. Participants reported using facial expressions strategically. Common emotional expressions included amusement, elation, confusion, interest, and surprise. Less frequent used expressions such as shame, fear or disgust were employed selectively, based on readiness and client need. The use of facial expressions was particularly important for clients working on emotional regulation, theory of mind, or pragmatic language. Clinicians described exaggerated expressions for younger children or autistic clients to increase salience. 

Beyond facial cues, nonverbal strategies reported included hand gestures, body posture, proximity shifts, and eye gaze to manage behavior, scaffold communication, and signal turn-taking. Gestures like pointing, modeling, or signing ``more" or ``stop" were especially common. These actions were adapted to suit each client's cognitive, sensory, and emotional profile.

When asked to reflect on SARs, participants saw opportunities for them to model emotions, assist in turn-taking, and provide consistent, low risk presence. Many participants believed SARs could help improve engagement, especially for children who were reluctant to interact with adults. Others saw potential in SARs facilitating emotion recognition and story-based role-playing. However, some cautioned that SARs must be used with care to avoid overstimulation or confusion in clients with sensory sensitivities. 

\subsubsection{Discussion}
A thematic analysis conducted on the posts by the participants focused on \textbf{Expression and Interpretation of Emotions in SLP Practice}.

\textit{\textbf{Emotional Ambiguity and Overlap:}} Participants frequently acknowledged that certain emotional states -- such as confusion, frustration, and embarrassment -- are harder to distinguish based on facial expressions alone. These emotions often share similar features or may be misinterpreted without additional context. Some illustrative quotes are:\\

\begin{tabular}{|p{10cm}}
\textit{"It is really difficult to produce a facial expression}\\
\textit{that would differentiate frustration, from anger,} \\
\textit{from disgust."}
\end{tabular}

\vspace{0.75em}

\begin{tabular}{|p{10cm}}
\textit{"Disappointment, sadness, and confusion can all be}\\
\textit{interpreted as anger, especially with younger}\\
\textit{kids who are on the autism spectrum."}
\end{tabular}\\

\textit{\textbf{Influence of Individual Variability and Masking:}} Participants emphasized that individuals differ in how naturally or intentionally they express emotions. Personality, neurodivergence, fatigue, distraction, or trauma can result in facial expressions that do not align with internal states.\\

\begin{tabular}{|p{10cm}}
\textit{"I tend to have about 100 things going at once...} \\
\textit{the facial expression that I made did not fit the} \\
\textit{situation."}
\end{tabular}

\vspace{0.75em}
\begin{tabular}{|p{10cm}}
\textit{"Some kids like to see that you are sad... for those}\\
\textit{kids, I would intentionally not make any facial}\\
\textit{expressions."}
\end{tabular} \\

\textit{\textbf{Cultural and Contextual Considerations:}} Respondents noted that cultural norms and situational context greatly influence how emotions are shown and interpreted. Cultural rules around emotional display (e.g., being reserved, overly polite) impact expressiveness and interpretation accuracy. \\

\begin{tabular}{|p{10cm}}
\textit{"Some cultures are more direct in expressing emotions}\\
\textit{like anger or sadness, while others may be more} \\
\textit{reserved and subtle."}
\end{tabular}

\textit{\textbf{Multimodal Emotional Interpretation:}} Participants stressed the need to rely on multiple communicative channels—including intonation, vocal quality, body language, and rhythm—to accurately interpret emotions. \\

\begin{tabular}{|p{10cm}}
\textit{"Intonation, rate, rhythm, stress, and muscular} \\
\textit{tension do a lot of heavy lifting."}
\end{tabular}

\vspace{0.75em}
\begin{tabular}{|p{10cm}}
\textit{"Facial expressions… sometimes I think we just use} \\
\textit{those for confirmation of what we’re already} \\
\textit{figuring out."}
\end{tabular} \\

\textit{\textbf{Implications for Clinical Practice:}} Participants shared strategies for supporting clients with social-pragmatic differences, including masking their own emotional expressions, using explicit instruction, and contextual teaching of emotions in therapy sessions.\\

\begin{tabular}{|p{10cm}}
\textit{"Facial expressions are tricky because they are} \\
\textit{usually something we can control and use} \\
\textit{intentionally."}
\end{tabular}

\vspace{0.75em}
\begin{tabular}{|p{10cm}}
\textit{"In speech therapy, facial expression can help} \\
\textit{the client understand what you are thinking and} \\
\textit{how you are feeling… but some emotions are harder}\\
\textit{to show nonverbally."}
\end{tabular} \\

\subsection{Perceived Roles and Benefits - Week 3}

\subsubsection{Activity}
Week 3 focused on envisioning roles for SARs in existing session strategies. Clinicians overwhelmingly saw SARs as useful in motivational and supportive roles. When asked about roles during intervention, SARs were frequently selected as companions (100\%), motivators (92\%), role-playing partners (77\%), and instructors (77\%). Other valued roles included therapist assistants (69\%), facilitator (46\%), and data collector (31\%).
In home-based practice, SARs were similarly viewed as companions (92\%) and motivators (83\%). Participants emphasized that SARs could help overcome the challenge of limited parental involvement or difficulty generalizing skills outside the clinic.
Some participants worried about the potential for a child to be overreliant on SARs or reduced interactions with clinicians. Others raised concerns, specifically regarding the possibility of SARs replacing SLPs to reduce labor costs. The participants stressed that SARs should remain tools to enhance-not replace-the clinician's presence, expertise, and relationship building capacity.

Participants were asked to choose anticipated benefits of using SARs during interventions. The results of these are listed: Increased engagement and motivation - 100\%, Novelty and interest - 92\%, Support with repetitive tasks - 67\%, Consistency in therapy delivery - 83\%, and Enhanced data collection and analysis - 25\%.

\subsubsection{Discussion}
A thematic analysis of Week 3's discussion post was conducted. The main theme of Week 3 was \textit{Integrating SARs into SLP Intervention}.

\textit{\textbf{Supplementary Role, Not Replacement:}} Many participants emphasized that SARs should serve as assistive tools rather than as replacements for clinicians. They expressed discomfort with SARs taking full control of sessions and saw the value in human-led therapy enhanced by robotic consistency and novelty.\\

\begin{tabular}{|p{10cm}}
\textit{"I would not want to fully 'turn over control'}  \\
\textit{of the session… but I could definitely see} \\
\textit{an assistant-type role."}
\end{tabular}

\vspace{0.75em}
\begin{tabular}{|p{10cm}}
\textit{"I wouldn’t feel comfortable with the robot} \\
\textit{taking lead in session but more as a} \\
\textit{tool and activity."}
\end{tabular} \\

\textit{\textbf{Enagement Through Novelty and Motivation:}} One of the most consistently mentioned benefits of SARs was their ability to motivate clients, particularly those who are disengaged, have repetitive therapy routines, or are drawn to technology.\\

\begin{tabular}{|p{10cm}}
\textit{"Therapy that is presented across [a tech]} \\
\textit{medium might have the sustained engagement}\\ 
\textit{factor."}
\end{tabular}

\vspace{0.75em}

\begin{tabular}{|p{10cm}}
\textit{"SARs could gamify boring tasks like speech} \\
\textit{sound practice."}
\end{tabular}\\

\textit{\textbf{Domain Specific Opportunities and Challenges:}} Participants explored SARs’ usefulness across therapy domains—articulation, language, social communication, and even stuttering—but highlighted different needs and limitations within each. These are described in the following:

\textbf{Articulation}
\begin{itemize}
    \item Need for visual modeling of articulators was noted as a major limitation.
    \item SARs could aid in massed practice, auditory bombardment, and data tracking.
\end{itemize}

\begin{figure}
    \centering
    \includegraphics[width=\linewidth]{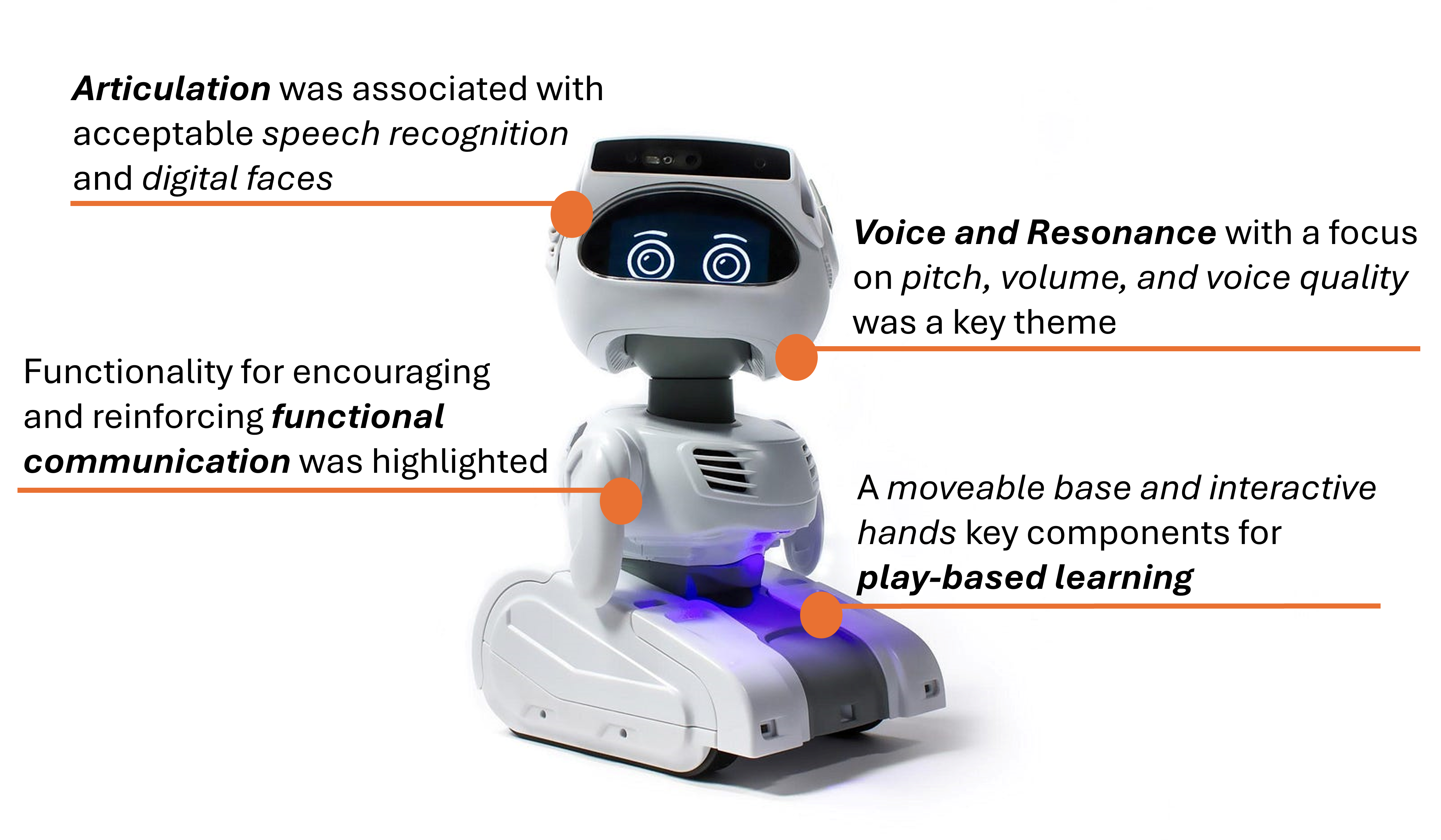}
    \caption{Using an existing embodiment solution like the Misty II~\cite{mistyrobotics}, we can map the main themes that participants considered when reflecting on possible functional requirements for SARs in clinical spaces.\vspace{-1em}}
    \label{fig:misty_summary}
\end{figure}

\textbf{Language}
\begin{itemize}
    \item SARs could be useful for question-asking, storytelling, and comprehension tasks.
\end{itemize}

\textbf{Social Communication}
\begin{itemize}
    \item SARs seen as potential role-playing partners for pragmatic skill-building
    \item Consistent programming could simulate social routines, then introduce unexpected responses to support generalization.
\end{itemize}

\textbf{Fluency (Stuttering)}
\begin{itemize}
    \item One participant asked: Could SARs offer safe spaces for stuttering, supporting stuttering-affirming therapy?\\
\end{itemize}

\begin{tabular}{|p{10cm}}
\textit{"SARs would be excellent role-play}\\
\textit{participants."}
\end{tabular}

\vspace{0.75em}

\begin{tabular}{|p{10cm}}
\textit{"They could be a word or story generator}\\
\textit{... and to do data collection."}
\end{tabular}\\

\textit{\textbf{Client-Specific Factors:}} Clinicians stressed the need to individualize SAR use based on client age, personality, diagnosis, sensory preferences, and tech familiarity. Some clients may find robots engaging, while others may be overstimulated, frightened, or distracted.\\

\begin{tabular}{|p{10cm}}
\textit{"Each client is different... some will be}\\
\textit{distracted, some may even be scared."}
\end{tabular}\\

\textit{\textbf{Technical and Ethical Considerations:}} Participants expressed curiosity and concern around SARs’ technical capacity and the ethics of implementation. Questions were raised about the capabilities of current SARs in speech recognition, linguistic analysis, and client data privacy.\\

\begin{tabular}{|p{10cm}}
\textit{"How are robots these days when it comes}\\
\textit{to detecting nuances of speech sound} \\
\textit{disorders?"}
\end{tabular}

\subsection{Feature Preferences by Clinical Domain - Week 4}

\subsubsection{Activity}
Week 4 focuses on different therapeutic domains and asked participants to identify key features that SARs should possess to effectively support intervention.

\begin{itemize}
    \item \textbf{Articulation}: Speech recognition and digital faces were deemed essential for providing accurate models and feedback. Participants highlighted the importance of real-time audio-visual feedback, integration of mouth modeling, and the ability to discriminate correct from incorrect productions.
    \item \textbf{Fluency}: Real-time data collection was highly valued, mainly for tracking disfluency frequency, type, and duration. Several participants suggested incorporating EMG or tension monitoring for clients working on stuttering modification. A few participants advocated for affirming, non-corrective feedback styles.
    \item \textbf{Voice and Resonance}: Feedback on pitch, volume, and vocal quality was highlighted as important, specifically for clients working on maintenance. SARs could provide reminders, track baseline voice parameters, and cue safe vocal behaviors.
    \item \textbf{Language (Expressive/Receptive)}: Participants envisioned SARs asking WH questions, modeling grammar, and supporting comprehension. Personalization, adaptable interaction, and child-friendly designs were emphasized. Digital faces, movable bases, and interactive hands were seen as useful for play-based learning. 
    \item \textbf{Hearing}: Participants stated that SARs could use captioning, sign language (via tactile limbs), and lip-reading support (via digital faces) to assist clients with hearing impairments. Accessibility features such as amplification and proximity responsiveness were also suggested. 
    \item \textbf{AAC \& Communication Modalities}: SARs were envisioned as AAC communication partners, modeling device use, prompting selections, and reinforcing functional communication. Some described SARs as avatars translating AAC inputs into lifelike actions.
    \item \textbf{Swallowing}: Though less familiar, some SLPs proposed SARs as motivational figures, visual models for feeding gesttures, or tools integrated with biofeedback to support safe swallowing and hydration habits.
    \item \textbf{Cognition}: For memory, attention, and problem-solving, participants suggested that SARs could help with reminders, sequencing tasks, and tracking progress. Participants emphasized the importance of flexible, accessible designs, and real-world applicability.
    \item \textbf{Social Communication}: SARs could act as emotional modeling tools, social partners, and pragmatic coaches. Facial expressions, conversational turn-taking, and role-play were frequently mentioned capabilities. However, participants urged that SARs should be used with care when applying them to clients sensitive to overload. 
\end{itemize}

\subsubsection{Discussion}
A thematic analysis of Week 4's discussion post was conducted. The main theme of Week 4 was \textit{Essential and Overlooked SAR Features for Clinical Use}.

\textbf{\textit{Accuracy and Precision in Speech and Language Processing:}} One of the most consistent themes was the need for SARs to accurately analyze speech, detect articulatory and phonological errors, and offer reliable feedback. Participants noted this as a non-negotiable for robots to be effective in articulation therapy and related speech tasks.\\

\begin{tabular}{|p{10cm}}
\textit{"The technology needs to be VERY precise}\\
\textit{to be functional for robot-client exchange."}
\end{tabular}

\vspace{0.75em}

\begin{tabular}{|p{10cm}}
\textit{"Can they identify phonological processes}\\
\textit{or articulatory errors/distortions?"}
\end{tabular}\\

\textbf{\textit{Role as Motivator, Companion, or Support Figure:}} Many saw SARs as most effective in motivational or supportive roles, especially for clients with low engagement or high anxiety. The idea of SARs offering low-stakes interaction and emotional encouragement was appealing across use cases. \\

\begin{tabular}{|p{10cm}}
\textit{"They could be great for motivating}\\
\textit{patients and building confidence."}
\end{tabular}

\vspace{0.75em}

\begin{tabular}{|p{10cm}}
\textit{"Great tools for post-stroke or }\\
\textit{cognition patients to have a friend..."}
\end{tabular}\\

\textbf{\textit{Customizability and Sensory Design:}} Participants expressed interest in SARs that were visually dynamic, adaptable in appearance, and sensitive to users’ sensory needs. The idea of robots that change colors, move fluidly, or adapt based on user preference emerged as a valuable engagement strategy.\\

\begin{tabular}{|p{10cm}}
\textit{"The ability to change the appearance of}\\
\textit{the device...like changing color or light scheme."}
\end{tabular}

\vspace{0.75em}

\begin{tabular}{|p{10cm}}
\textit{"Motion... a static device would probably}\\
\textit{not be as interesting to most younger users."}
\end{tabular}\\

\textbf{\textit{Data Collection and Learning Over Time:}} Participants highlighted the value of SARs capturing session data, including speech accuracy, support levels, and interaction patterns. Several envisioned SARs that could learn from users over time to deliver more personalized and adaptive interactions.\\

\begin{tabular}{|p{10cm}}
\textit{"Document precise data on the child's accuracy}\\
\textit{and level of support required."}
\end{tabular}

\vspace{0.75em}

\begin{tabular}{|p{10cm}}
\textit{"A program that learns from the experiences} \\
\textit{with its user over time."}
\end{tabular}\\

\textbf{\textit{Limitations in Complex Pragmatics and Contextual Nuance:}} Clinicians questioned SARs' ability to manage higher-level social and pragmatic tasks, such as interpreting sarcasm, emotional undertones, or abstract reasoning. This concern reflects current limitations in AI understanding of subtle human interaction.\\

\begin{tabular}{|p{10cm}}
\textit{"Social cognition is so complex... no single prod}\\
\textit{or poke tends to make a difference."}
\end{tabular}

\vspace{0.75em}

\begin{tabular}{|p{10cm}}
\textit{"I'm skeptical about how accurate they would be}\\
\textit{able to track data...so many variables."}
\end{tabular}\\

\textbf{\textit{Safety and Accessibility Innovations:}} One participant suggested innovative health-related features, such as vital sign monitoring or fall detection, which go beyond speech therapy into broader caregiving and wellness applications.\\

\begin{tabular}{|p{10cm}}
\textit{"If these SARs...could detect when signs of}\\
\textit{life have stopped and could call for help."}
\end{tabular}

\section{Discussion}
SAR4SLPs provided a nuanced exploration into how speech-language pathologists perceive and envision SARs across the range of therapy contexts. The Slack platform yielded in-depth responses that participants could reflect on asynchronously. With this insight into the quality of data that the ARC method can elicit, applying this strategy in other HRI co-design contexts could further improve access and acceptability when designing SAR systems.

Results from Week 1 to Week 4 indicate a cautious optimism and a need for clarity, personalization, and clinical alignment along the development and integration of SARs. Participants expressed moderate comfort with technology and their open-mindedness to using new tools. However, the majority of the participants noted their limited knowledge and familiarity. Aligning with work from Heerink et al and Kennedy et al this suggests that exposure and informed collaboration is critical when adopting technology into healthcare~\cite{Heerink2010, kennedy2017impact}. After being introduced to examples of SARs, SLP participants were able to identify possible applications like client motivation, consistent cueing, and modeling social interactions. 

Considering expressive capabilities, the responses from the SLPs indicate a strong connection in current SLP expressive use and expressive capacities of SARs. Participants underscored the importance of exaggeration, active communication, and tailoring to the individual needs of clients related to their development and sensory profile. This is particularly relevant to developing a SAR, as it suggests that dynamic and controllable expression can support therapy rapport and comprehension. 

SLPs also focused on the varied responses to SARs depending on their experiences. Clients might find SARs threatening while others could be distracted or overwhelmed. Results echo prior work that raised concerns within personalization research, reinforcing adaptive and user-aware systems~\cite{irfan_23}.

SARs were also seen as potentially viable for articulation and language therapy, particularly for drill-based repetition, data tracking, and vocabulary modeling. However, participants noted that they were more skeptical about SARs supporting fluency or emotional skills where subtle human context is necessary for appropriate responses. This also mirrors work from Dautenhahn et al showing that human-like presence is insufficient without context and emotional intelligence~\cite{dautenhahn2007socially}.

In Week 4, responses revealed that SLPs had strong opinions about the necessary design features for SARs for therapy interventions. While mobility and physical resemblance were less important, SLPs prioritized speech recognition accuracy, custom feedback, and expression. These features point to SAR development that should focus on functional transparency and cognitive alignment rather than surface level imitation. 

SAR4SLP contributes important practitioner-informed perspectives to the field of socially assistive robotics. It reaffirms that successful integration of SARs into SLP contexts depends not only on technological sophistication, but also on the tools' ability to engage, adapt, and respect the therapeutic alliance. 

\section{Limitations and Future Works}
Several limitations of this study should be acknowledged. First, the sample size was relatively small (n=13), which limits generalizability. All participants were practicing SLPs, primarily in school-based or pediatric settings, which may not capture the full range of therapy environments. Additionally, responses were self-reported and may be subject to social desirability bias or limitations in recall.

Future work should aim to expand participant diversity across settings (e.g., hospitals, private practice), populations (e.g., adults, multilingual clients), and geographic regions. Longitudinal studies could explore how perceptions and usage of SARs evolve over time and in response to training. Finally, co-design studies that directly involve SLPs and clients in the development and testing of SARs will be critical to ensure relevance, usability, and ethical implementation.

\section*{Acknowledgment}

This material is based upon work supported under the AI Research Institutes program by National Science Foundation and the Institute of Education Sciences, U.S. Department of Education through Award \# 2229873 - AI Institute for Transforming Education for Children with Speech and Language Processing Challenges. Any opinions, findings and conclusions or recommendations expressed in this material are those of the author(s) and do not necessarily reflect the views of the National Science Foundation, the Institute of Education Sciences, or the U.S. Department of Education. This work was supported by the National Science Foundation (IIS
2150394)
This work was supported by Dr. Julie Kientz and the Computing for Healthy Living and Learning Lab.

\bibliographystyle{IEEEtran}
\bibliography{conference_101719}

\end{document}